\documentclass[journal]{IEEEtran}

\IEEEoverridecommandlockouts                              

\usepackage{wrapfig}
\usepackage{algpseudocode}
\usepackage{times}
\usepackage{float}
\usepackage{epsfig}
\usepackage{graphicx}
\usepackage{algorithm}
\usepackage{subcaption}
\usepackage{array}
\usepackage{amsmath}
\usepackage{amssymb}
\usepackage{balance}
\usepackage[hidelinks]{hyperref}
\usepackage{booktabs}
\newcolumntype{P}[1]{>{\centering\arraybackslash}p{#1}}

\usepackage[flushleft]{threeparttable}
\usepackage{xcolor}
\title{\LARGE \bf
Where Shall I Touch? Vision-Guided Tactile Poking for Transparent Object Grasping}

\author{Jiaqi Jiang$^{1,2}$, Guanqun Cao$^{1}$, Aaron Butterworth$^{1}$, Thanh-Toan Do$^{3}$ and Shan Luo$^{1,2}$ 
\thanks{Manuscript received: 29th December, 2021; revised: 9th May, 2022 and 12th July 2022; accepted 30th July, 2022).}
\thanks{This work was funded in part by the EPSRC ViTac project (EP/T033517/1), and in part by the University of Liverpool and China Scholarship Council Award.}
\thanks{$^{1}$J. Jiang, G. Cao, A. Butterworth and S. Luo are with the smARTLab, Department of Computer Science, University of Liverpool, Liverpool L69 3BX, United Kingdom. E-mails: \tt\small\{jiaqi.jiang, g.cao, a.butterworth, shan.luo\}@liverpool.ac.uk.}%
  \thanks{$^{2}$J. Jiang and S. Luo are also with Department of Engineering, King's College London, London WC2R 2LS, United Kingdom. E-mail: {\tt\small \{jiaqi.1.jiang, shan.luo\}@kcl.ac.uk.}}
  \thanks{$^{3}$T.-T. Do is with Department of Data Science and AI, Monash University, Clayton, VIC 3800, Australia. E-mail: {\tt\small toan.do@monash.edu}.}

}

\begin{document}

\maketitle

\begin{abstract}
Picking up transparent objects is still a challenging task for robots. The visual properties of transparent objects such as reflection and refraction make the current grasping methods that rely on camera sensing fail to detect and localise them. However, humans can handle the transparent object well by first observing its coarse profile and then poking an area of interest to get a fine profile for grasping. 
Inspired by this, we propose a novel framework of vision-guided tactile poking for transparent objects grasping. In the proposed framework, a segmentation network is first used to predict the horizontal upper regions named as \textit{poking regions}, where the robot can poke the object to obtain a good tactile reading while leading to minimal disturbance to the object's state. A poke is then performed with a high-resolution GelSight tactile sensor. Given the local profiles improved with the tactile reading, a heuristic grasp is planned for grasping the transparent object. 
To mitigate the limitations of real-world data collection and labelling for transparent objects, a large-scale realistic synthetic dataset was constructed.
Extensive experiments demonstrate that our proposed segmentation network can predict the potential poking region with a high mean Average Precision (mAP) of 0.360, and the vision-guided tactile poking can enhance the grasping success rate significantly from 38.9\% to 85.2\%.
Thanks to its simplicity, our proposed approach could also be adopted by other force or tactile sensors and could be used for grasping of other challenging objects. 
All the materials used in this paper are available at \url{https://sites.google.com/view/tactilepoking}.
\end{abstract}

\begin{IEEEkeywords}
Transparent objects, tactile sensing, visual perception, multi-modal sensing, object segmentation, robot grasping and manipulation.
\end{IEEEkeywords}

\section{INTRODUCTION}
\IEEEPARstart{T}{ransparent} objects are widely used in our daily life, e.g., glass cups, plastic bottles and glass pan lids in a kitchen. They are also common in research laboratories~\cite{eppel2020computer,burger2020mobile}, e.g., vials, glass flasks and Petri dishes. Many of these objects are fragile and easy to break, therefore need to be handled with extra attention. To have robots work in such environments, it is essential for robots to have safe interaction with transparent objects. Without this capability, transparent objects may be broken or have their contents spilled. Broken glass or spilled liquid will pose hazards to the robot and people that share its space.
\begin{figure}[t]
  \includegraphics[width=\linewidth]{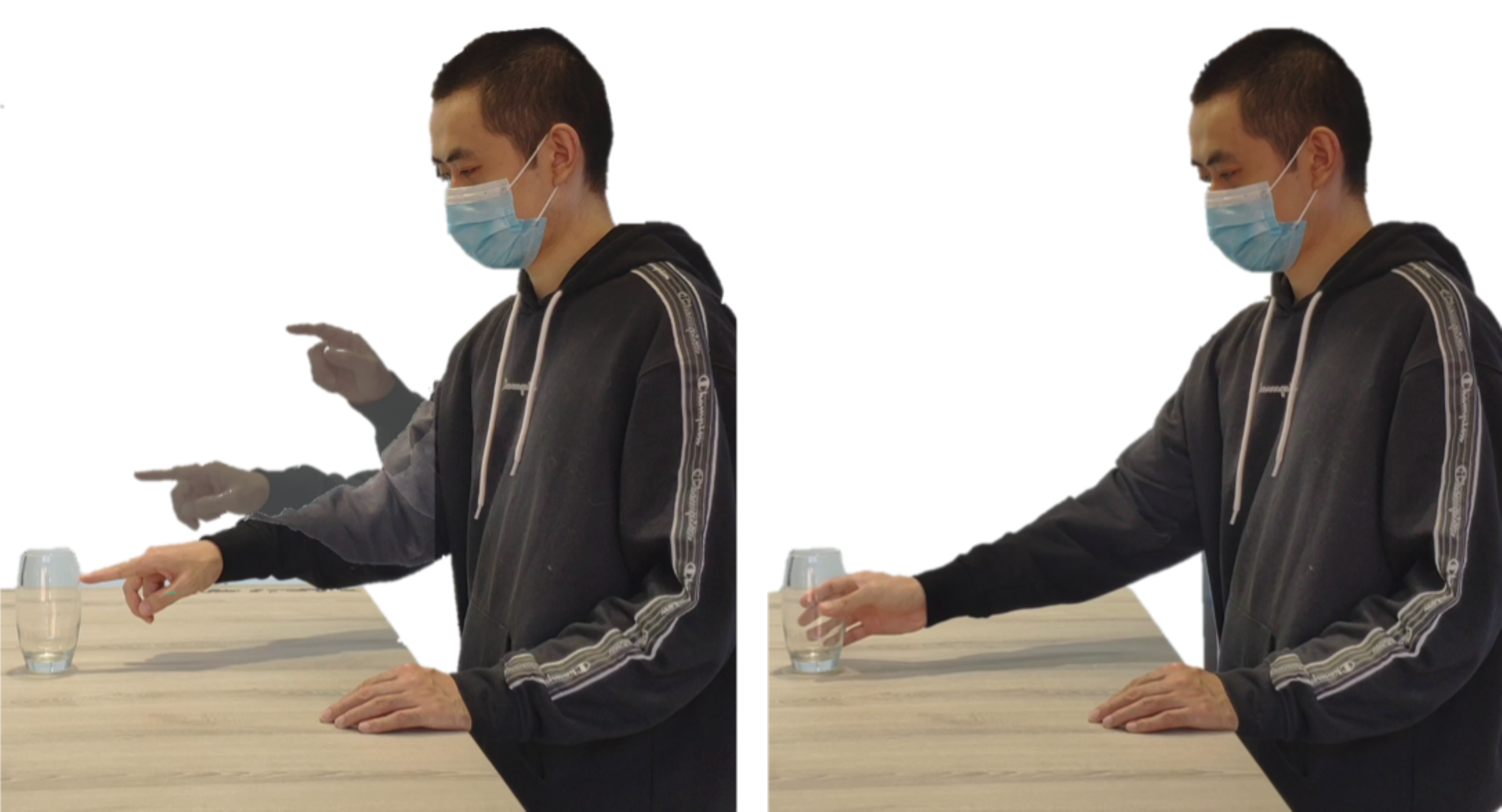}
   \caption{An illustration of vision-guided tactile poking for transparent object grasping as we do in our daily lives. As shown in the figure on the left, a glass cup on the table is hard to detect due to its transparency, which brings difficulties to grasping the cup. Before we grasp the cup, we first have a glance at the cup and predict potential areas for contact. We then move our hands following the vision guidance to poke the cup, as shown on the left. The contact with the cup will give us an accurate localisation of the cup, which facilitates a stable grasp of the cup, as shown on the right.}
\label{fig:intuition}
\end{figure}

However, it is still challenging for a robot to detect and grasp transparent objects \cite{sajjan2020clear,weng2020multi}. Most objects in previous object detection and grasping research have been opaque and the perception of transparent objects remains a challenging problem. Compared to opaque objects, transparent objects lack salient features in their surfaces such as colour and texture features. Moreover, their transparent materials violate the Lambertian assumption that optical 3D sensors (e.g., LiDAR and RGB-D cameras) are based on: the opaque objects reflect light evenly in all directions, resulting in a uniform surface brightness from all viewing angles, however, the surfaces of transparent objects both reflect and refract light. Hence, most of the depth data of transparent objects from depth sensors is invalid or contains unpredictable noise. Due to these challenges, most of the current grasping methods that rely on accurate depth information from cameras cannot be directly applied to the grasping of transparent objects.

Humans grasp objects with rich sensory information~\cite{johansson2009coding,lederman2009haptic}, such as the visual information obtained from eyes and the tactile feeling via physical interaction. It is common that vision with a wide field of view is used first for fast localisation of objects, then touch providing accurate perception of compliance and contact force is used to align hand posture or grip strength to enable a stable grasp~\cite{jenmalm2000visual}. 
Research on the coordination of human visual and tactual input \cite{bower1970coordination,barrett2008infants} has shown that we use vision to anticipate an object's physical characteristics prior to contact, preparing the hand for grasping, whereas touch takes over control after the object is in the hand, in particular while interacting with transparent objects~\cite{sheya2010development} as shown in Fig.~\ref{fig:intuition}.

Inspired by those observations, we propose a novel vision-guided tactile poking approach for grasping transparent objects in this paper. Different from most prior studies~\cite{sajjan2020clear, weng2020multi} using only RGB-D images to address these challenges of transparent objects, our method integrates visual and tactile sensing so that they aid each other and improve the grasping performance.
We first train a deep neural network named \textit{PokePreNet} with synthetic RGB images to predict the poking regions that are with similar surface normals to the table surface. The contacts with those areas contribute to good tactile readings while leading to minimal disturbance to the state of the object. 
A robotic arm equipped with a tactile sensor is then guided to contact those regions, so as to generate informative local profiles of the contacted transparent objects. Finally, using the improved profiles, a heuristic grasp proposal is generated for grasping the transparent object.

To evaluate the performance of our PokePreNet, we construct a high-quality synthetic dataset as well as a real-world test benchmark with over 9,000 RGB images and their corresponding ground truth annotations. To bridge the gap between the simulation and the real world, we randomise the simulator to expose the model to a wide range of environments while training. 
Our experiments demonstrate that our proposed method can learn vision-guided tactile poking regions, with a high mean Average Precision (mAP) of 0.360, and generalise to transparent objects in the real world. 
We also conduct real robot experiments and results show that our proposed method can enhance the success rate of transparent object grasping from 38.9\% to 85.2\%, compared to a vision based grasping. 
Thanks to its simplicity, the proposed method can be adapted to other settings that use other force or tactile sensors, and can also be used for grasping of other challenging objects.

Our contributions can be summarised as follows:
\begin{itemize}
    \item We propose a vision-guided tactile poking approach for grasping transparent objects, which is the first of its kind;
    
    \item We introduce a novel poking region segmentation network trained with a pixel-level Positive-Negative-balanced loss, which boosts segmentation performance;
    
    \item We have collected a high-quality synthetic dataset for transparent object perception and grasping to bridge the reality gap, which is the largest of its kind.  
\end{itemize}

The rest of this paper is structured as follows. Section~\ref{Section:2} reviews the related works and Section~\ref{Section:3} introduces the robot setup and the dataset; Section~\ref{Section:4} details our proposed vision-guided tactile poking method for transparent object grasping; Section~\ref{Section:5} analyses the experimental results; Finally, Section~\ref{Section:6} summarises the paper and discusses the work.

\section{Related Works} \label{Section:2}

\subsection{Transparent Object Grasping}

There are two types of methods for robots to perform transparent object grasping in the literature. The first aims to reconstruct depth maps of transparent objects, so as to mitigate the sensor failures in depth images. In~\cite{klank2011transparent}, an approach was proposed that matches pixels from Time-of-Flight images first and then reconstructs an approximated surface with triangulating methods. To enhance the matching speed and reduce the influence of noise, a method was proposed in~\cite{phillips2016seeing} to match transparent edges instead of all pixels. However, those methods require multiple views of one object, which is not suitable for the case when the camera is fixed. To address this challenge, some other studies~\cite{sajjan2020clear, zhu2021rgb, jiang2022a4t} focus on reconstructing the missing or noisy depth regions of transparent objects using a single RGB-D image. In~\cite{sajjan2020clear}, a global optimisation algorithm was adopted to reconstruct the depth values that are removed based on predicted object masks. In~\cite{zhu2021rgb}, a local implicit neural representation built on ray-voxel pairs was proposed to reconstruct depth information incorporated with an iterative self-correcting refinement model. In~\cite{jiang2022a4t}, an affordance-based depth reconstruction framework was proposed to facilitate the robotic manipulation of transparent objects.

Rather than reconstructing a depth map, \cite{weng2020multi, xu20206dof} generate the grasp proposal with only RGB images or noisy depth maps as input. In~\cite{weng2020multi}, transfer learning was used to transfer the grasping model trained on depth maps to transparent object grasping with RGB images. In~\cite{xu20206dof}, a two-stage approach was proposed to estimate 6-DoF pose of transparent objects from a single RGB-D image, which can be used to assist transparent object grasping.
Nonetheless, there has been no work on transparent object grasping with both visual and tactile information. To our best knowledge, this is the first work to achieve the task.

\subsection{Object Grasping using Vision and Touch} 
The coordination between vision and touch sensing plays an important role in robot perception and has been applied to a number of different tasks~\cite{luo2017robotic} such as object recognition~\cite{luo2018vitac, lee2019touching}, shape exploration~\cite{luo2015localizing}, and object grasping~\cite{calandra2018more, cui2020grasp}. 
The visual-tactile features can be fused via direct concatenation~\cite{calandra2018more} or a Self-Attention mechanism~\cite{cui2020self}. Moreover, the coordination of vision and touch allows us to develop regrasping policies that will best grasp the object~\cite{calandra2018more}. However, the above studies either assume that the object position is known or use the depth information from camera to localise the object. These assumptions are not suitable for detecting and grasping transparent objects that are placed at a random location on the table due to their noisy or missing depth maps. In contrast, in this work, we use visual feedback to provide geometric cues for guiding the tactile sensor to contact the transparent object, which facilitates its grasp.

\subsection{Sim2Real Learning for Transparent Objects}
Synthetic datasets have been used in a wide array of applications, such as object segmentation~\cite{nanbo2020learning}, human pose estimation~\cite{chen2016synthesizing}, and tactile object classification~\cite{gomes2021generation}. 
However, there are only a few synthetic datasets for transparent objects. Most of those datasets~\cite{chen2018tom, zhu2021rgb} were generated without considering the subtle effect of transparent objects, e.g., specular highlights and caustics. In contrast, our simulation method can not only generate realistic images of transparent objects with such effects considered, but also provide detailed annotations, using the LuxCoreRender~\cite{LuxCoreRender} engine. Compared to the rendering method used in~\cite{sajjan2020clear} that uses Cycles~\cite{Cycles} engine, our method generates more natural synthetic images and builds the glass shader in an easier way.

\begin{figure}[t]
\begin{center}
\includegraphics[width=\linewidth]{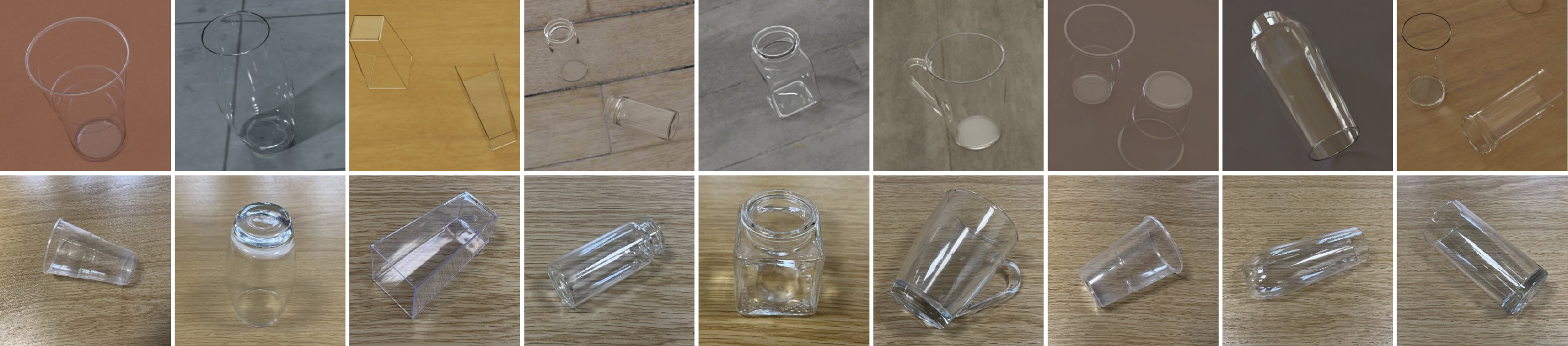}

\end{center}
  \caption{There are 9 objects in both synthetic (the first row) and real-world datasets (the second row), from left to right: large disposable cup, highball cup, rectangular cup, vial, jar, mug, small disposable cup, champagne cup and tumble cup.  }
\label{fig:syn_obj}
\end{figure}

\section{System Setup}\label{Section:3}
In this section, we first introduce a high-quality synthetic dataset generation framework. It can auto-generate the poking areas that are hard to be annotated by humans, e.g., the side surface of a cylindrical cup. Then we introduce the robot setup used for the real world data collection and experiments. We introduce the system setup before the methodology as we believe the synthetic data generation will also be a contribution of this paper, and can help the reader better understand the methods to be introduced later.

\subsection{Synthetic Data Generation}
We first generate the synthetic data of transparent objects, i.e., RGB images, depth images, surface normals and instance masks, before we conduct real world experiments, due to two reasons. First, key cues that determine the poking regions, i.e., surface normals, cannot be obtained in real world experiments. Second, labels like instance masks of transparent objects can be generated automatically in synthetic data, whereas it is challenging and time consuming for human annotators to annotate instance masks in real data of transparent objects.

We use Blender’s physics engine~\cite{Blender} and LuxCoreRender rendering engine~\cite{LuxCoreRender} to generate our synthetic dataset. Through simulating the flow of light, LuxCoreRender can not only produce photo-realistic images, but also simulate important effects caused by the presence of transparent objects such as reflections and caustics. The dataset consists of 9 objects modelled after real-world transparent glass objects, as shown in Fig.~\ref{fig:syn_obj}. To enrich the variety of the synthetic data, we employed 33 HDRI lighting environments and 20 textures for the ground plane underneath the transparent objects.

To bridge the gap between simulation and real environments, we first set the camera intrinsics based on the parameters of the Intel RealSense D415 camera that we use in the real experiments. Then we randomly select one HDRI lighting environment and one ground plane surface texture applied with a random rotation angle for each scene. Finally, several CAD model objects were created above the plane surface to increase the learning efficiency. For each scene, the ground truth data from Blender includes: (1) rendered monocular RGB image, (2) aligned depth in meters, (3) instance masks of all transparent objects, (4) the camera pose, and (5) surface normals of the scene.

\begin{figure}[t]
   \includegraphics[width=\linewidth]{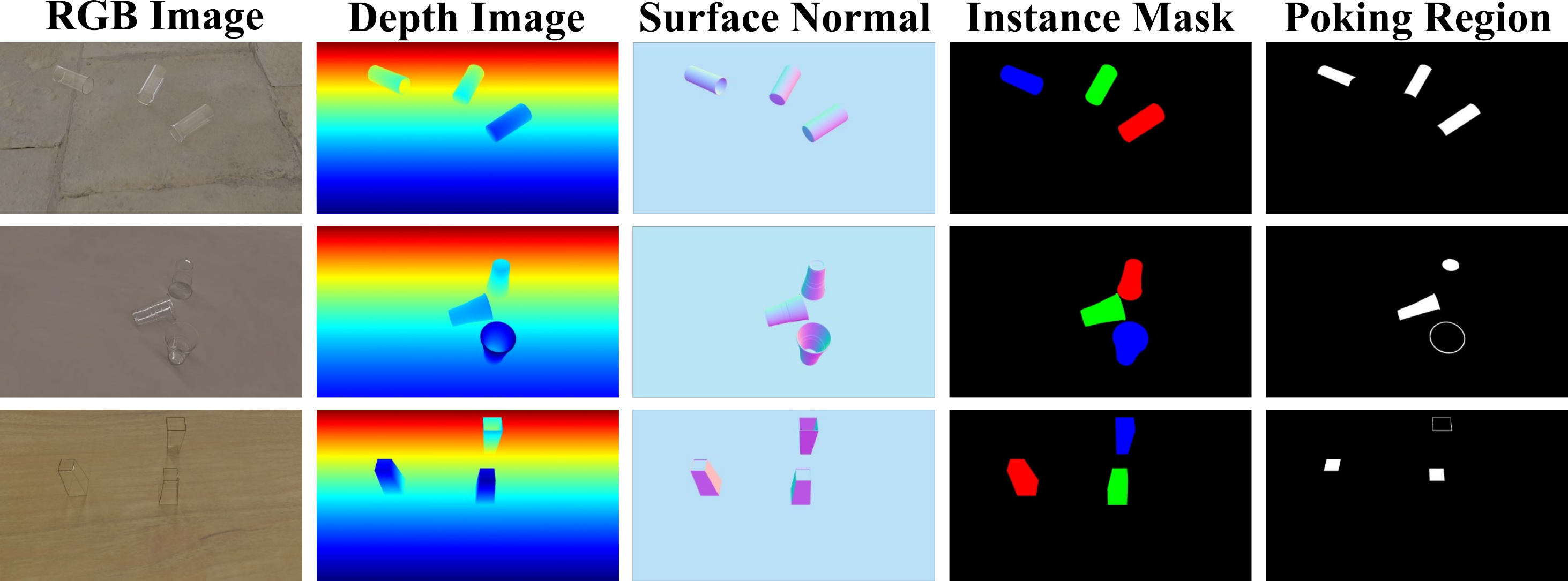}
   \caption{Three visualisation examples of our synthetic dataset. The first four columns are RGB images, depth images, surface normals and instance masks of three scenes with a few transparent objects rendered in Blender, respectively, and the poking region masks in the last column are generated from them.}
\label{fig:synthetic data}
\end{figure}

\begin{figure}[t]
   \includegraphics[width=\linewidth]{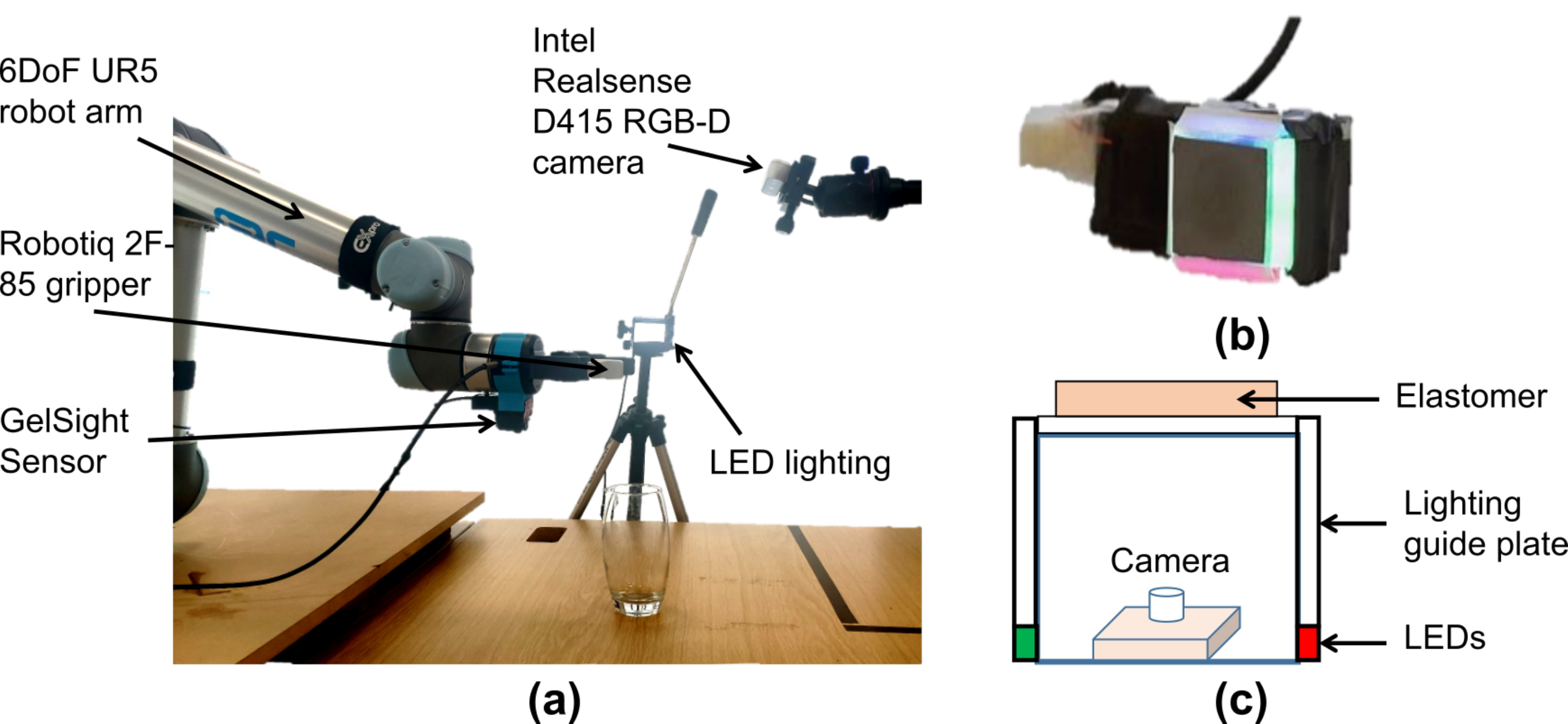}
   \caption{\textbf{Robot setup.} \textbf{(a)} An overview of the experimental setup that consists of a UR5 robotic arm, Robotiq 2F-85 gripper, a GelSight sensor and an Intel RealSense D415;
   \textbf{(b)} The GelSight Sensor; \textbf{(c)} A sketch diagram of the GelSight sensor.}
\label{fig:robotsetup}
\end{figure}

Using the rendered data, the ground truth of poking regions can be generated as follows. First, we get the dot product map via calculating the dot product of each pixel and the table surface normal. Then, we apply a pre-defined threshold to the dot product map to get initial poking regions. However, not all the initial poking regions are suitable for tactile poking, for example, the inner surface of a cup. To remove those areas, we calculate the height map relative to the ground plane using the depth image and the camera pose. If the height of one pixel is lower than a predefined threshold, the pixel will not be set as part of the poking region. 
Figure \ref{fig:synthetic data} shows some examples of rendered images and their corresponding ground truth of poking regions for transparent objects. In total, there are over 9,000 views of 9 objects generated in the sythetic dataset.

\begin{figure*}[t]
\begin{center}
  \includegraphics[width=0.9\linewidth]{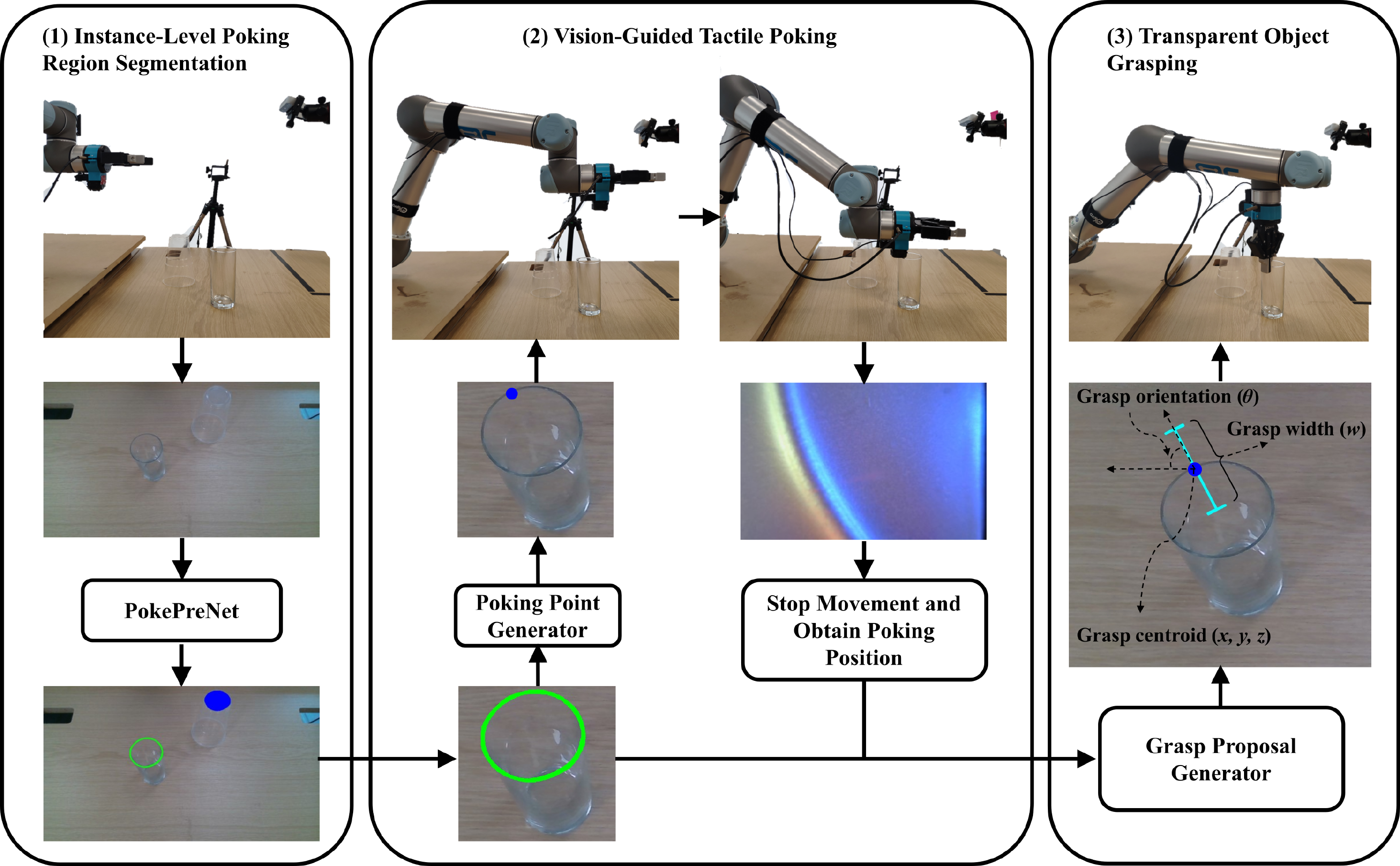}

\end{center}
   \caption{An overview of our vision-guided tactile poking approach for transparent object grasping. 
   \textbf{From left to right:} 
   First, the PokePreNet takes the RGB image, and outputs the segmented poking regions where different colours represent different instances. Then based on the detected poking regions, the poking point generator is used to generate the potential poking point that guides the robotic arm to move towards the transparent object until the equipped GelSight sensor contacts the object. Lastly, with predicted poking region and the obtained local profiles from tactile poking, a heuristic grasp proposal is generated for grasping the transparent object.}
\label{fig:overview}
\end{figure*}
\subsection{Robot Setup for Transparent Object Grasping}
\label{Section:RobotSetup}
As shown in Fig. \ref{fig:robotsetup}, our robot setup consists of a 6-DOF UR5 robot equipped with a Robotiq 2F-85 adaptive gripper and a GelSight sensor, as well as an Intel RealSense D415 RGB-D camera mounted on the tripod for overseeing the environment. The GelSight is a camera-based optical tactile sensor that can detect contact and capture fine details of the object surface. In a GelSight sensor, a webcam is placed under an elastomer that captures the deformations of the elastomer on the top when contacted. The sensor has a flat sensing area of $14 mm \times 10.5 mm$ and can capture tactile images with a resolution of $1280 \times 720$ at a frequency of 30 Hz \cite{cao2020spatio}. In previous works~\cite{calandra2018more}, the gripper's fingers were replaced with GelSight sensors to generate tactile images for grasped objects. However, this replacement limits the sensing area to the inner side of the finger. To address this limitation, as shown in Fig.~\ref{fig:robotsetup}(a), the GelSight sensor is attached to the end-effector’s side with a 90° angle to achieve a tactile poking action, i.e., the GelSight's y-axis and z-axis coincide with the end-effector's y-axis and x-axis, respectively. The rotation between the end-effector and the GelSight sensor is then set as $(0,\frac{\pi}{2},0)$, and the translation is obtained with an Opti-Track Motion Capture system. Moreover, the classic Tsai hand-eye calibration method~\cite{tsai1989new} is applied to the calibration between the RealSense camera and the UR5 robot.

\subsection{Real-World Dataset Collection}
To test the generalisation of the proposed PokePreNet, we also create a dataset of real-world transparent objects in the laboratory space. As illustrated in Fig.~\ref{fig:syn_obj}, there are 4 transparent plastic and 5 glass objects in both our synthetic and real-world datasets.
The real-world dataset consists of 180 images of those nine known objects used in synthetic training data. Each image contains one object randomly placed on the table. Due to the difficulty of annotating the side surface of cylindrical objects, only the rectangular cup and the jar, i.e., the third column and fifth column in Fig.~\ref{fig:syn_obj}, have the cases where they stand on their sides.

\section{Methodology}\label{Section:4}
In this work, we propose a vision-guided tactile poking approach for transparent object grasping, with an overview of the framework illustrated in Fig. \ref{fig:overview}. First, the poking region prediction network (\textit{PokePreNet}) takes a single RGB image and outputs the poking region segmentation in the instance level. Based on the detected poking region, a poking point is then generated to guide the robotic arm to move towards the transparent object. The robotic arm will stop once a contact between the equipped GelSight sensor and the object is detected. Finally, with the predicted poking region and obtained local profiles from tactile poking, a heuristic grasp proposal is generated for grasping the transparent object.

\subsection{Poking Region Segmentation} \label{sec3::A}
The poking region segmentation is treated as an instance segmentation problem. In the instance segmentation, every pixel will be simultaneously classified whether it belongs to the poking region and which instance it is part of. 
One of the most popular instance segmentation techniques is Mask R-CNN~\cite{he2017mask}. However, the poking region only occupies a small part of the bounding box, which causes a bad precision of Mask R-CNN.
To solve this issue, our PokePreNet introduces two novel improvements to the original Mask R-CNN for segmenting the poking regions: (1) a larger output feature map via adding more deconvolutional layers; (2) a new pixel-level Positive-Negative-balanced loss. 

\noindent \textbf{Larger output feature map}. We add two more deconvolutional layers to increase the size of poking region masks from $28\times28$ to $112\times112$. The filters in all the deconvolutional layers have a size $S_{f}$ of $2\times2$, with zero padding $d=0$ and stride $s=2$, which can double the size of the feature map:
\begin{equation} \label{equ1}
S_{o}=s *\left(S_{i}-1\right)+S_{f}-2 * d
\end{equation}
where $S_i$ and $S_o$ are the sizes of the input feature map and the output feature map, respectively.

\noindent \textbf{Pixel-level Positive-Negative-balanced loss}. We use a multi-task loss $L$ to jointly train the instance segmentation network to predict the object class, bounding box position, and poking region mask on each Region of Interest (RoI) as follows:
\begin{equation}
    \label{equation::1}
    L=L_{cls}+L_{loc}+L_{mask}
\end{equation}
where $L_{cls}$ is the multinomial cross entropy loss; $L_{loc}$ is the \textit{Smooth L1} loss~\cite{ren2015faster} between the regressed box offsets $t=\{t_x, t_y, t_w, t_h\}$ and the ground-truth box offsets $v=\{v_x, v_y, v_w, v_h\}$: 
\begin{equation}
    L_{loc}(t,v) = \sum\limits_{k\in\{x,y,w,h\}}{Smooth_{L1}(t_k-v_k)} 
\end{equation}
where 
\begin{equation}
{Smooth}_{L_{1}}(x)= \begin{cases}0.5 x^{2} & \text { if }|x|<1 \\ |x|-0.5 & \text { otherwise.}\end{cases}
\end{equation}

\noindent $L_{mask}$ is the poking region mask loss. Following~\cite{he2017mask, ren2015faster}, the weighting factors for each sub-loss are set to 1.

\begin{figure}[t]
  \centering
  \includegraphics[width=0.9\linewidth]{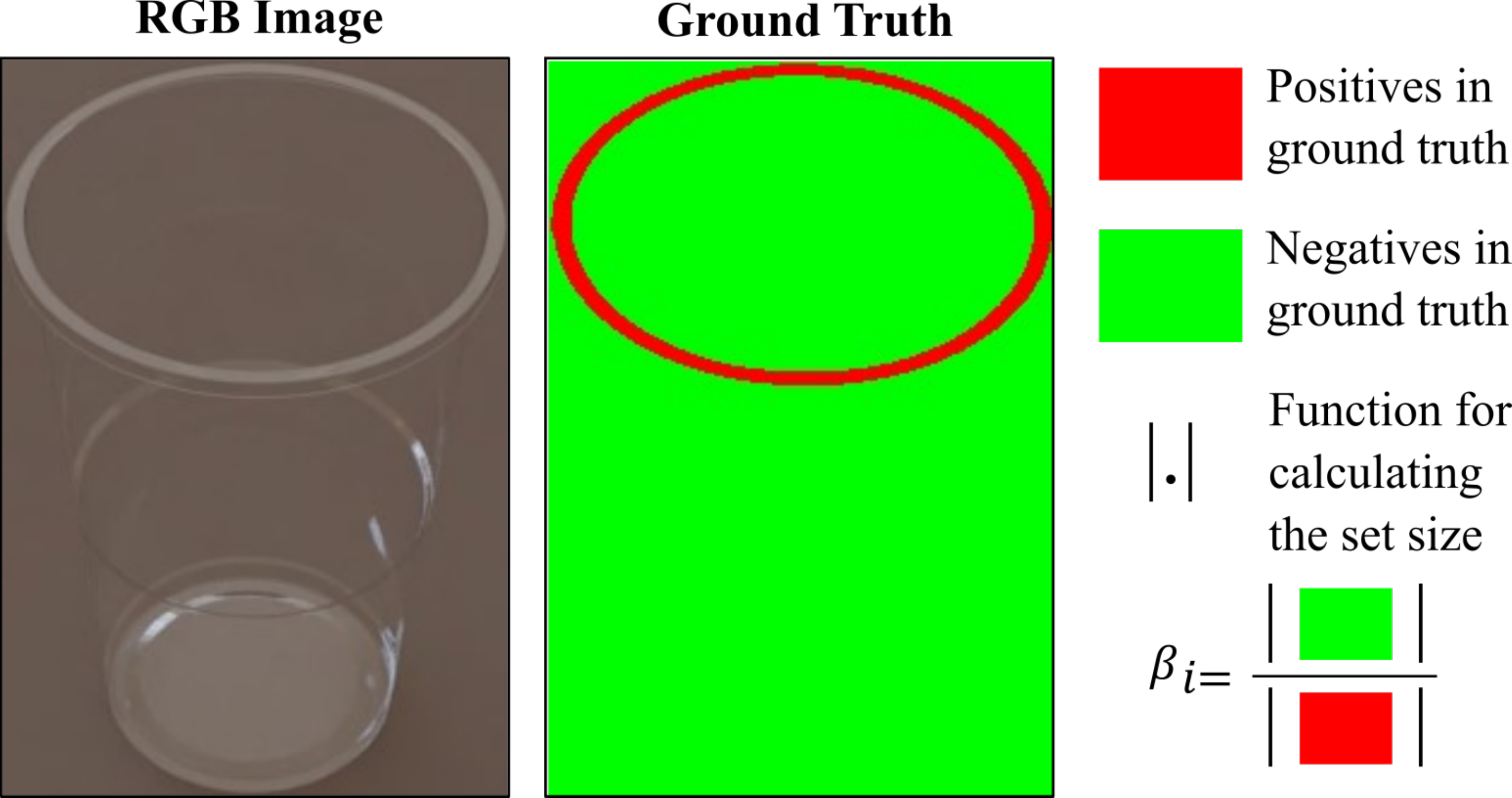}
  \caption{An illustration of how the Positive-Negative-balanced weight $\beta_i$ is computed. Red pixels in the ground truth are the positive pixels that are part of the poking region, whereas green pixels are the negative pixels that are not part of the poking region. }
\label{fig:sketch}
\end{figure}

In the vanilla Mask R-CNN, the average binary cross-entropy loss is used for training instance masks. However, the distribution of positive/negative pixels (positive pixels are the pixels that are part of the poking regions, and negative pixels are ones that are not) in the objects such as the cup in Fig.~\ref{fig:sketch} is heavily biased: only 5\% of the bounding box area is part of the poking region. To this end, the cross-entropy loss from the poking region only contributes to a small part of the total loss, and leads to a bad precision of the poking region. 

To address this issue, we define the following pixel-level Positive-Negative-balanced (PN) loss for the poking region mask $L_{mask}$ in Eq. \ref{equation::1}:
\begin{equation}
\begin{array}{r}
L_{mask}(X_{i})=-\beta_{i}\sum\limits_{j \in Y^{+}_{i}} \log \operatorname{Pr}\left(y_{j}=1\right|X_{i})\\
-\sum\limits_{j\in Y^{-}_{i}} \log \operatorname{Pr}\left(y_{j}=0\right)|X_{i})
\end{array}
\end{equation}
where $Y^{+}_{i}$ and $Y^{-}_{i}$ denote the positive and negative ground truth label sets for the $i^{th}$ RoI $X_{i}$, respectively; $\beta_i$ is the weight on an instance basis to balance the loss between positive and negative pixels, as illustrated in Fig.~\ref{fig:sketch}.

Specifically, $\beta_i$ is set to ${|Y^{-}_{i}|}/{|Y^{+}_{i}|}$ and 1 when $|Y^{+}_{i}|$ is larger than 0 and equal to 0, respectively. $|.|$ function is used for calculating the set size, and $j$ represents the pixel index.  $\operatorname{Pr}\left(y_{j}=1\right|X_{i}) = \sigma\left(a_j\right)\in[0,1]$ is computed using sigmoid function $\sigma\left(.\right)$ on the activation value $a_j$ at pixel $j$.

In our initial experiments, we find that the PN loss boosts the poking region segmentation performance in the experiments in the real environment. However, in our synthetic dataset results, there are some extremely small poking regions and as a result $\beta_i$ is very large in such cases, which results into a large number of false positives and lowers the performance.
Hence, to enhance the performance in the simulation, we use a log function to restrict large values and use a Log-Positive-Negative-balanced (LPN) loss for $L_{mask}$ instead with $\beta_i$:

\begin{equation}
\beta_{i} = \left\{
\begin{matrix}
\ln(\frac{|Y^{-}_{i}|}{|Y^{+}_{i}|}) && \text{if} \quad |Y^{+}_{i}|>0 \\
1&& \text{if}\quad |Y^{+}_{i}|=0
\end{matrix}
\right
.
\end{equation}

The proposed method can be recognised as a kind of Hard Example Mining method~\cite{shrivastava2016training}, i.e., mining of examples that are hard to be classified or detected. The hard examples in this work are the pixels from the instance with a small poking region where a biased pixel distribution exists. Hence, we use this heuristic to accelerate the mining of hard examples without the need to classify each pixel in each RoI individually, which makes our method more efficient.

\subsection{Vision-guided Tactile Poking}
\label{Section:VisionGuidedTactilePoking}
Given the detected poking region from Sec.~\ref{sec3::A}, we generate a poking point~$\boldsymbol{P_{t}}=[x_t, y_t]$ in the image frame for every transparent object. To generate the poking point, we first find the external contour of the poking region mask using OpenCV function \textit{findContours}. Then we use OpenCV function \textit{fitEllipse} to fit the contour and get the centroid $\boldsymbol{P_c}$. Similar to~\cite{lin2020using}, the poking points are generated based on the primitive shapes. As shown in the output of our PokePreNet in Fig.~\ref{fig:overview}, the 2-D poking regions are simplified into two types of primitive shapes: a simply connected mask (the blue mask) if $\boldsymbol{P_c}$ is part of the poking region, and a ring shape connected mask (the green mask) if $\boldsymbol{P_c}$ is out of the poking region. 

If the poking region is a simply connected mask, the poking point will be set to the ellipse centroid $\boldsymbol{P_c}$, as centroids are widely used for grasping the objects with simple rectangular or cylindrical shapes~\cite{asif2018ensemblenet}.
On the other hand, if the poking region is a ring shape mask, $\boldsymbol{P_c}$'s nearest positive pixel will be set as the poking point to avoid getting the GelSight sensor into the object.
The algorithm used to find the poking point in the image frame is summarised in Alg. \ref{alg::1}.

\begin{algorithm}
\caption{Poking point generation} \label{alg::1}
\begin{algorithmic}[1]
\Require $M_{poking}$: a poking region mask.
\Ensure $\boldsymbol{P_{t}}=[x_t, y_t]$: a poking point in the image frame
\State external\_contour $\gets$ findContours($M_{poking}$)
\State ellipse $\gets$ fitEllipse(external\_contour)
\If{ellipse.centroid in $M_{poking}$}
    \State $[x_t, y_t] \gets$ ellipse.centroid

\Else
    \State $[x_t, y_t] \gets$ findNearestPositive(ellipse.centroid)

\EndIf
\end{algorithmic}
\end{algorithm}

Guided by the poking point, the robotic arm is moved towards the transparent object until the equipped GelSight sensor contacts the object. The GelSight sensor is set parallel to the table so as to minimise the horizontal force and avoid the change of the object's state.
The tactile contact is detected with a simple image subtraction-based algorithm~\cite{gomes2020geltip}.
First, a tactile image is captured as the reference. Then, the element-wise absolute difference between the reference and the current frame is computed and applied with a binary thresholding in every channel. Finally, the contact will be recognised in the current frame, if the number of positive pixels in the difference frame is larger than a predefined threshold.
We also considered detecting the contact by thresholding the Structural Similarity Index Measure (SSIM) of the reference and contact images. However, compared to the image subtraction-based algorithm, the computational cost of SSIM is much higher (0.2s vs. 0.02s for processing each tactile image).
To stop the robotic arm as soon as a contact is detected and avoid destroying fragile transparent objects, the image-subtraction method is used.

\subsection{Heuristic Transparent Object Grasping} \label{sec3::C}
Based on the predicted poking region and the object's local profiles (i.e., contact position) from the tactile poking, a heuristic grasp representation in the world frame is generated for the top-down parallel grasping. The grasp representation is defined as a 5-dimensional vector $\boldsymbol{G_{hrst}} = [x, y, z, w, \theta]$ as shown in Fig.~\ref{fig:overview}, where $[x, y, z]$ represents the grasp centroid in the world frame. $w$ and $\theta$ represent the width and the orientation of the heuristic grasp, respectively. Note that $\theta$ is the one-dimensional angle around the vertical axis of gravity direction to facilitate the top-down parallel grasping.

If $\boldsymbol{P_c}$ belongs to the poking region, the poking position $\boldsymbol{P^W_t}$ in the world frame will be equal to the position of centre $\boldsymbol{P^W_c}$. Hence, a centroid-based grasp~\cite{asif2018ensemblenet} is used for grasping the transparent object. In detail, $[x, y, z]$ of $\boldsymbol{G_{hrst}}$ will be set to $\boldsymbol{P^W_t}$. The grasp width $w$ and the orientation $\theta$ are set to the maximum value of gripper width, and the fitted ellipse rotation angle for grasping along the short axis of the ellipse, respectively.  
If $\boldsymbol{P_c}$ is not part of the poking region, the grasp centroid will be set according to the distance $D(\boldsymbol{P^W_c},\boldsymbol{P^W_t})$ between $\boldsymbol{P{_c}}$ and $\boldsymbol{P{_t}}$ in the world frame. Under the assumption that $\boldsymbol{P{^W_c}}$ and $\boldsymbol{P{^W_t}}$ are at the same height, the centroid of the fitted ellipse in the world frame $\boldsymbol{P{^W_c}}$ can be calculated with a pin-hole camera model.

If $D$ is larger than the half of the finger width, the gripper finger could be inserted into the transparent object. Hence, an edge grasp is used for grasping the transparent object as the grasp proposal shown in Fig.~\ref{fig:overview}. 
$[x, y, z]$ will be the poking position $\boldsymbol{P^W_t}$. The grasp width $w$ and the orientation $\theta$ are set to the twice of $D$ and parallel to the vector $<\boldsymbol{P^W_c},\boldsymbol{P^W_t}>$, respectively. 
Otherwise, a centroid-based grasp is used and the grasp position will be set to $\boldsymbol{P{^W_c}}$. The grasp width $w$ and orientation $\theta$ are set to the maximum value and the fitted ellipse rotation angle. The algorithm used to generate the heuristic grasp is summarised in Alg. \ref{alg::2}.

\begin{algorithm}
\caption{Heuristic grasp generation} \label{alg::2}
\begin{algorithmic}[1]
\Require $\boldsymbol{P_t^W}$: poking position in the world frame; 
$M_{poking}$: a poking region mask;
ellipse: fitted ellipse from Alg. \ref{alg::1}.
\Ensure  $\boldsymbol{G_{hrst}} = [x, y, z, w, \theta]$: a heuristic grasp proposal.
\If{ellipse.centroid in $M_{poking}$} \Comment{centroid grasp}
    \State $[x, y, z] \gets$ $\boldsymbol{P_c^W} \gets \boldsymbol{P_t^W}$ 
    \State $w \gets$ maximum\_gripper\_width
    \State $\theta \gets$ ellipse.rotation\_angle
\Else
    \State $\boldsymbol{P_c^W} \gets$ calculateWorldPosition(ellipse.centroid)
    \State $D \gets$ calculateDistance($\boldsymbol{P_c^W},\boldsymbol{P_t^W}$)
    \State $Angle \gets$ calculateAngle($\boldsymbol{P_c^W},\boldsymbol{P_t^W}$)
    \If{ $D > 0.5 \times$finger\_width} \Comment{edge grasp}
        \State $[x, y, z] \gets$ $\boldsymbol{P_t^W}$ 
        \State $w \gets$ 2$\times{D}$
        \State $\theta \gets Angle $
    \Else \Comment{centroid grasp}
        \State $[x, y, z] \gets \boldsymbol{P_c^W}$ 
        \State $w \gets$ maximum\_gripper\_width
        \State $\theta \gets$ ellipse.rotation\_angle
    \EndIf
\EndIf
\end{algorithmic}
\end{algorithm}

\begin{figure*}[h]
\centering
  \includegraphics[width=0.95\linewidth]{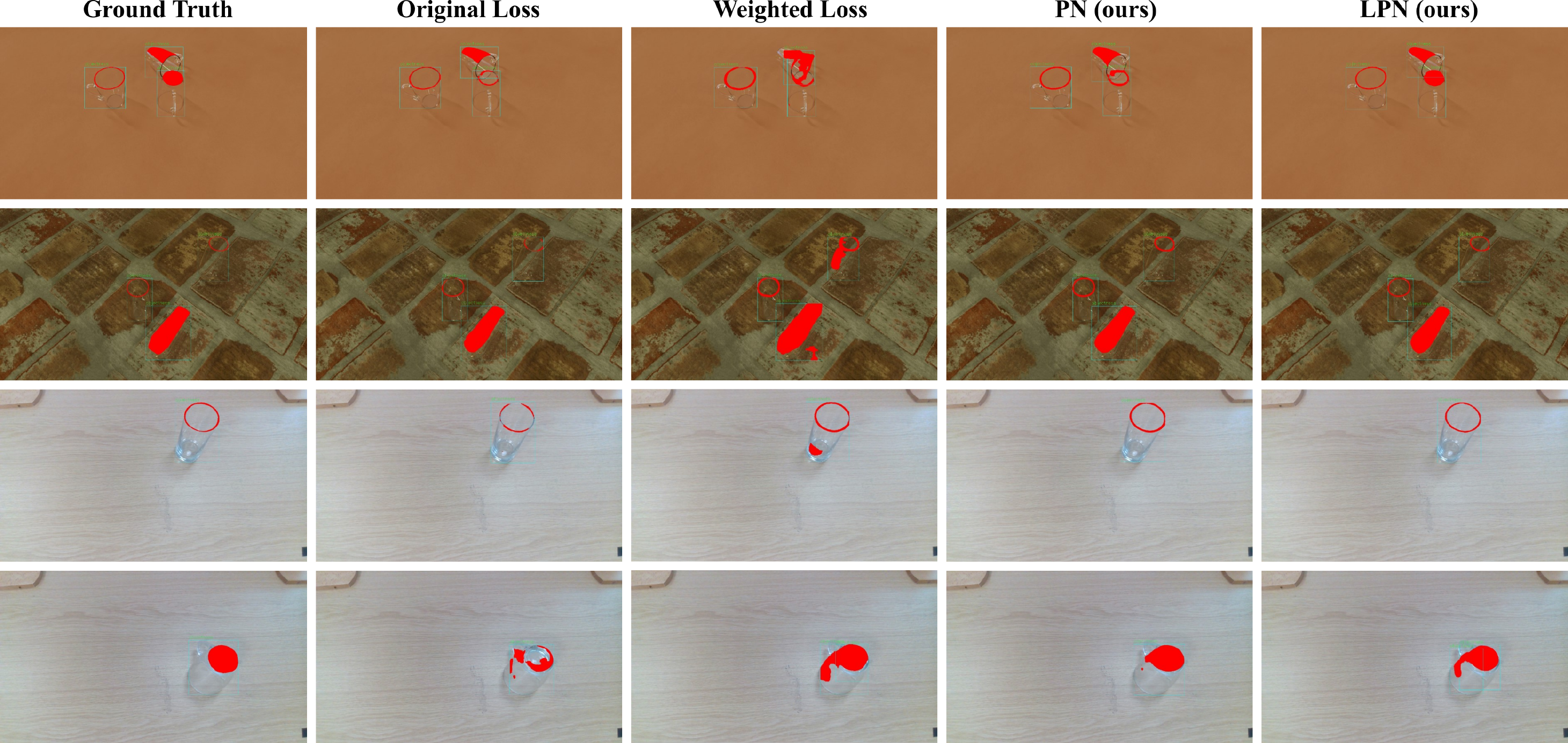}

   \caption{Visual comparison of poking region segmentation results using different loss functions. The top two rows and the bottom two rows compare results on the synthetic dataset and the real-world dataset, respectively. As shown above, our PN and LPN based methods generate much better poking regions, compared to the vanilla loss and weighted loss.}
\label{fig:losscomparison}
\end{figure*}

\section{Experiments}\label{Section:5}
In this section, we conduct a series of experiments to evaluate our vision-guided tactile poking for transparent objects grasping. The goal of the experiments are three-fold: (1) To evaluate the poking region segmentation accuracy of our PokePreNet in both synthetic and real-world datasets; (2) To investigate how poking regions improve the success rate of tactile poking against bounding boxes and instance masks; (3) To investigate how the feedback from tactile poking can improve the success rate of transparent objects grasping.

\subsection{Poking Region Segmentation Experiments}
To evaluate the poking region segmentation accuracy, the standard Average Precision (AP) metric is used. To be more specific, we used mean AP (mAP), AP$_{50}$ and AP$_{75}$, i.e., AP at different Intersection over Union (IoU) thresholds 50\% and 75\%, and AP$_{S}$, AP$_{M}$ and AP$_{L}$ (AP at different scales, i.e., small, medium and large). It should be noted that the object scale is determined by the poking region size instead of the bounding box size. We only evaluate the poking region segmentation results in this work, as the bounding box detection is not related to our tactile poking approach. Similar to the previous studies~\cite{he2017mask}, our PokePreNet uses a Region Proposal Network to extract 1000 proposals for each image. Our poking region segmentation experiments are organised as follows. Firstly, we compare our Positive-Negative-balanced loss (PN) and Log-Positive-Negative-balanced loss (LPN) against vanilla cross-entropy loss and weighted cross-entropy loss in both the synthetic and real-world datasets. Secondly, we analyse the effect of the output size of the poking region map. Thirdly, we examine the domain randomisation's effect on generalisation. 

\noindent \textbf{Evaluation of different loss functions.} We evaluate PokePreNet on both the synthetic and real-world benchmarks. Table~\ref{tab:re1} compares the performance of using different types of losses for poking region segmentation. The vanilla loss represents the average binary cross-entropy loss used in the vanilla Mask R-CNN. Weighted loss adds a fixed large weight to the cross-entropy loss of positive pixels.

In the synthetic dataset, the weighted loss and the PN loss although bring 7.9\% and 3.5\% gains in terms of AP${_{S}}$ respectively, nevertheless lead to a significant drop of the overall performance. This is because when the poking region areas are extremely small in the synthetic dataset, the balanced weight on positive pixels will result in more false positives and lower the performance. After using the log function to compress the value range of the balanced weight in PN loss, the LPN loss achieves an improvement of 8.7\% and 3.2\% on AP${_{S}}$ and mAP, respectively. 

In the real-world test benchmark, both our PN loss and LPN loss outperform the original loss and the weighted loss: PN loss leads to the best overall performance and LPN loss results in largest improvement on AP${_{M}}$.
These different trends might be caused by two main reasons: (1) Manual annotations of the real-world benchmark is not as accurate as synthetic annotations; (2) Domain randomisation is not sufficient to bridge the domain gap between the simulation and the real world. To address this problem, domain adaptation will be considered in the future work. We also show the qualitative results of poking region segmentation in Fig.~\ref{fig:losscomparison}.
As illustrated, the original loss and the weighted loss result in a lot of false negatives and false positives. Our PN loss and LPN loss yield highest quality poking region segmentation results on the real-world images and the synthetic images.

\begin{table}[h]
	\centering
		\caption{BASELINE COMPARISONS ON SYNTHETIC AND REAL BENCHMARK.}
		\label{tab:re1}
        \scalebox{0.9}{
		\begin{tabular}{c| c | c | c | c| c|c |c }
			\hline
			Test data &Loss type & mAP & AP$_{50}$ & AP$_{75}$ &AP$_{S}$ & AP$_{M}$ & AP$_{L}$ \\
			\hline
			\hline
    		Synthetic & vanilla  & 0.530 & 0.843& 0.600&0.020 & \textbf{0.509} & 0.744\\
    		Synthetic & weighted   & 0.468 & 0.865 & 0.467 & 0.099 & 0.399 & 0.733\\
    		Synthetic & PN [ours]   &0.472  & 0.749 & 0.500 & 0.053& 0.388 & 0.752 \\
    		Synthetic & LPN [ours]  & \textbf{0.562} & \textbf{0.916} & \textbf{0.601} & \textbf{0.107} & 0.507 & \textbf{0.775}\\ 
    		\hline
    		Real & vanilla  & 0.319 & 0.672& 0.248 & N/A & 0.149&0.540\\
    		Real & weighted   & 0.330 & 0.669 & 0.292&N/A &0.155&0.542\\
    		Real & PN [ours]  & \textbf{0.360}  & \textbf{0.778} & \textbf{0.304} & N/A& 0.181 & \textbf{0.576} \\
    		
    		Real & LPN [ours]  & 0.356 & 0.757 & 0.221 & N/A  & \textbf{0.234} & 0.536 \\ 
    		\hline
		\end{tabular}}
\end{table}

\begin{figure*}[t]
\centering
  \includegraphics[width=0.95\linewidth]{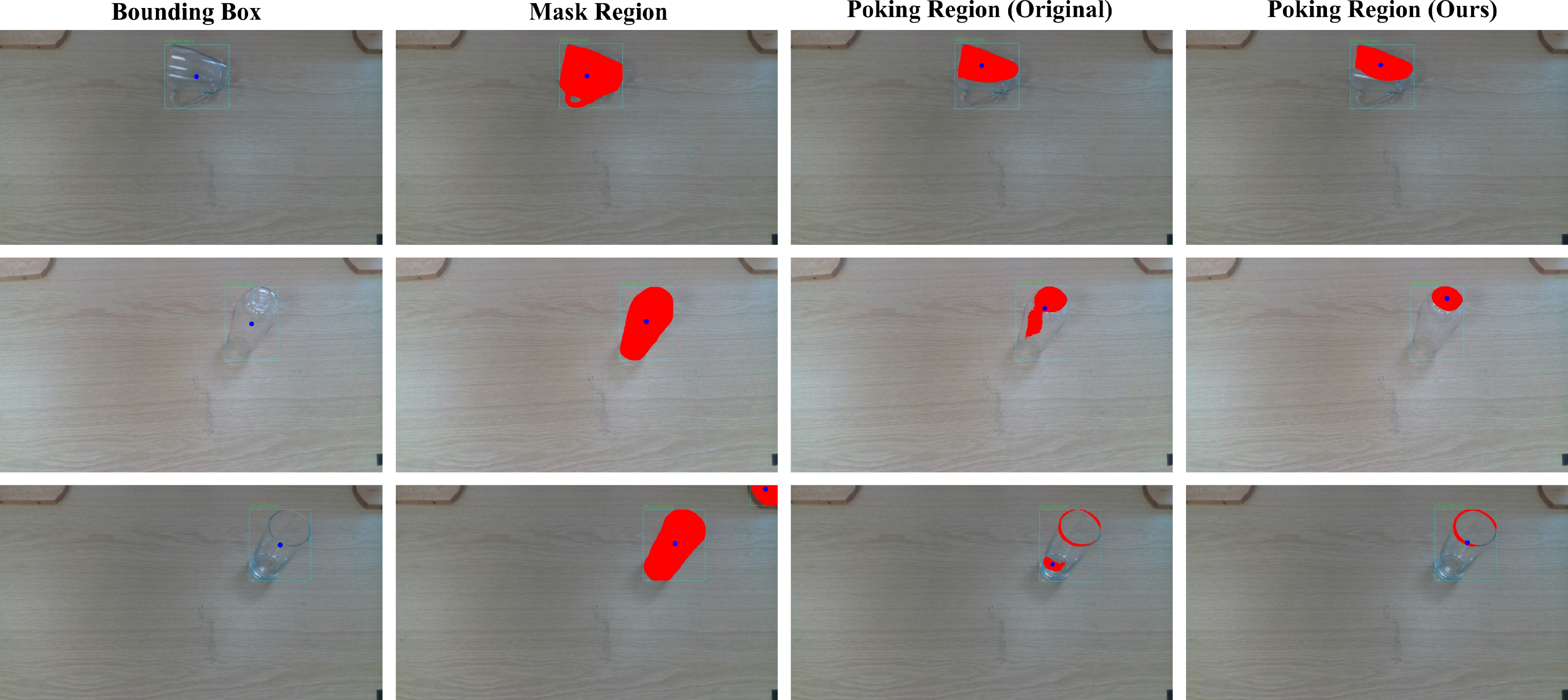}
   \caption{Examples of the poking points generated with the bounding box, the instance mask, the poking region with vanilla Mask R-CNN loss (Original) and the poking region with our PN loss (Ours). The red colour and blue dot represent the segmentation results and generated poking points, respectively.
}
\label{fig:touch_illustration}
\end{figure*}

\noindent \textbf{Evaluation of the poking region output feature map size.}
We also analyse the effect of the poking region output feature map size. Mask R-CNN uses only one deconvolutional layer to create the $28\times28$ mask map from the $14\times14$ feature map. Following the setup of Mask R-CNN, we add one, two, and three more deconvolutional layers to create the $56\times56$, $112\times112$, and $224\times224$ poking region mask map, respectively. 

Table~\ref{tab:re2} summarises the segmentation accuracy of the mentioned networks on both the synthetic dataset and real-world dataset. The results show that the segmentation accuracy is gradually improved when the output size is increased from $28\times28$ to $112\times112$, however, it is not further improved when adding one more deconvolutional layer to make the output size to $224\times224$, on both synthetic and real datasets. It should be noted that similar results have also been reported in~\cite{zhang2021refinemask}.
One possible reason is that $112\times112$ is large enough to show the details of the object and, compared to $224\times224$, is closer to the real feature map size of transparent objects. Moreover, the best models tested in the synthetic dataset and the real dataset are trained with our LPN loss and PN loss, respectively.

\begin{table}[h]
	\centering
		\caption{EFFECT OF OUTPUT MASK SIZE.}
		\label{tab:re2}
        \scalebox{0.9}{
		\begin{tabular}{c| c | c | c | c |c |c |c }
			\hline
			Test data & Output size & mAP  & AP$_{50}$ & AP$_{75}$ &AP$_{S}$ & AP$_{M}$ & AP$_{L}$   \\
			\hline
			\hline
    		Synthetic &$28\times28$   & 0.397  & 0.654  & 0.425 & 0.018  & 0.299 & 0.660 \\
    		Synthetic &$56\times56$    & 0.442 & 0.708 & 0.477 & 0.051 & 0.354  & 0.752  \\
    		Synthetic &$112\times112$  & \textbf{0.562} & \textbf{0.916} & \textbf{0.601} & \textbf{0.107} & \textbf{0.507} & \textbf{0.775}  \\
			Synthetic &$224\times224$   & 0.501  & 0.868 & 0.501  & 0.064 & 0.398 & 0.749 \\
    		\hline
    		Real &$28\times28$   &   0.198 & 0.425 & 0.183 &  N/A   & 0.081  & 0.337 \\
    		Real &$56\times56$     & 0.271 & 0.555 & 0.224 &  N/A   & 0.117  & 0.454  \\
    		Real &$112\times112$   & \textbf{0.360}  & \textbf{0.778} & \textbf{0.304} & N/A& \textbf{0.181} & \textbf{0.576} \\
			Real &$224\times224$   & 0.337  & 0.751 & 0.281 & N/A& 0.170 & 0.554 \\
    		\hline
		\end{tabular}}
\end{table}

\noindent \textbf{Evaluation of domain randomisation.}
Despite not being trained on real transparent objects for poking region segmentation, our models can be adapted well to the real-world domain. To evaluate the importance of our data generation methodology, we assessed the model's sensitivity to the number of unique textures seen in the training. Table~\ref{tab:re3} shows that the domain randomisation method via applying different textures significantly improves the mAP of poking region segmentation accuracy in real-world dataset from 31.3\% to 36.0\%. 

\begin{table}[h]
	\centering
		\caption{EFFECT OF DOMAIN RANDOMISATION (DR).}
		\label{tab:re3}
        \scalebox{1}{
		\begin{tabular}{c| c | c | c | c |c |c |c }
			\hline
			Test data & DR & mAP  & AP$_{50}$ & AP$_{75}$ &AP$_{S}$ & AP$_{M}$ & AP$_{L}$   \\
			\hline
			\hline
		Synthetic &$\times$    & 0.472 & 0.803 & 0.515 & 0.073 & 0.420  & 0.672  \\
    		Synthetic & $\surd$   & \textbf{0.562} & \textbf{0.916} & \textbf{0.601} & \textbf{0.107} & \textbf{0.507} & \textbf{0.775}\\

    		\hline
    			Real & $\times$   & 0.313 & 0.697 & 0.249 & N/A  & 0.160 & 0.506 \\
    		Real & $\surd$   & \textbf{0.360}  & \textbf{0.778} & \textbf{0.304} & N/A& \textbf{0.181} & \textbf{0.576} \\
    	    		\hline

		\end{tabular}}
\end{table}

\subsection{Vision-Guided Tactile Poking Experiments} 

In this subsection, we conduct real-world experiments to evaluate the performance of our vision-guided tactile poking method. 
We define that the poking action will be recognised as successful, if no protective stop happens to the robotic arm, and a contact happens between the transparent object and the GelSight sensor. During the poking motion, the robotic arm will stop if the height of the end-effector is lower than a predefined threshold, which means the poke misses the object entirely and will be taken as a failure.

Table~\ref{tab:re4} compares the success rates of vision-guided tactile poking methods using four different input sources for poking point generation. The ``bounding box" and ``mask region" approaches guide the tactile poking with the centroid position of predicted bounding boxes and predicted instance masks, respectively. 
The ``poking region" represents those methods that use the poking region segmentation results as the input of poking point generator. [Vanilla] and [Ours] respectively represent the PokePreNet trained with the vanilla binary cross-entropy loss and our PN loss. Similar to~\cite{sajjan2020clear}, we had 12 attempts in grasping of each object, i.e., a total of 108 attempts for testing the above approaches. The results show that the poking region is a better cue for guiding the tactile poking compared to bounding box and instance mask, and the better poking region segmentation contributed by our PN loss can further improve the poking success rate from 84.3\% to 89.8\%.

The poking points generated with different methods have been visualised in Fig.~\ref{fig:touch_illustration}.
It is noticed that the generated poking points based on bounding box and instance mask sometimes have different surface normals against the table surface (e.g., the first and second columns), which will lead to failed poking motions. We can also observe that the bad poking region segmentation caused by the vanilla cross-entropy loss can also result in a failed tactile poking as shown in the third column.


\begin{table}[h]
	\centering
		\caption{COMPARISON ON VISION-GUIDED TACTILE POKING.}
		\label{tab:re4}
        \scalebox{1}{
		\begin{tabular}{c| c|c|c|c  }
			\hline
			Object Category & BBox & Mask & PR (Vanilla) & PR (Ours)  \\
			\hline
			\hline
            Big disposable cup & 4/12 & 4/12 & 8/12 & 8/12 \\ \hline
            Highball cup & 5/12 & 5/12 & 9/12 & 11/12 \\ \hline
            Rectangular cup & 7/12 & 6/12 & 10/12 & 10/12 \\ \hline
            Vial & 8/12 & 8/12 & 12/12 & 12/12\\ \hline
            Jar & 12/12 & 12/12 & 12/12 & 12/12\\ \hline
            Mug & 6/12 & 8/12 & 10/12 & 12/12\\ \hline
            Small disposable cup & 5/12 & 6/12 & 8/12 & 8/12\\ \hline
            Champagne cup & 5/12 & 6/12 & 11/12 & 12/12\\ \hline
            Tumble cup & 6/12 & 7/12 & 11/12 & 12/12\\ \hline
            Average success rate & 53.7\% & 57.4\% & 84.3\% & 89.8\% \\
            \hline
		\end{tabular}}
\end{table}
\subsection{Transparent Object Grasping Experiments}
To demonstrate the advantage of vision-guided tactile poking, we compare four different objects grasping approaches, as shown in Table~\ref{tab:re5}. 
``Baseline1" and ``Baseline2" generate grasp proposals from object instance masks and poking regions (PR) based on the depth obtained from an RGB-D camera, respectively. In contrast, our methods use the contact position from tactile poking to generate the heuristic grasp proposal. Same as the tactile poking experiment, every grasping approach was tested with 12 attempts on each object including 4 attempts with the object upright, 4 attempts with the object upside down, and 4 attempts with the object standing on its sides.

\begin{figure}[t]
  \includegraphics[width=\linewidth]{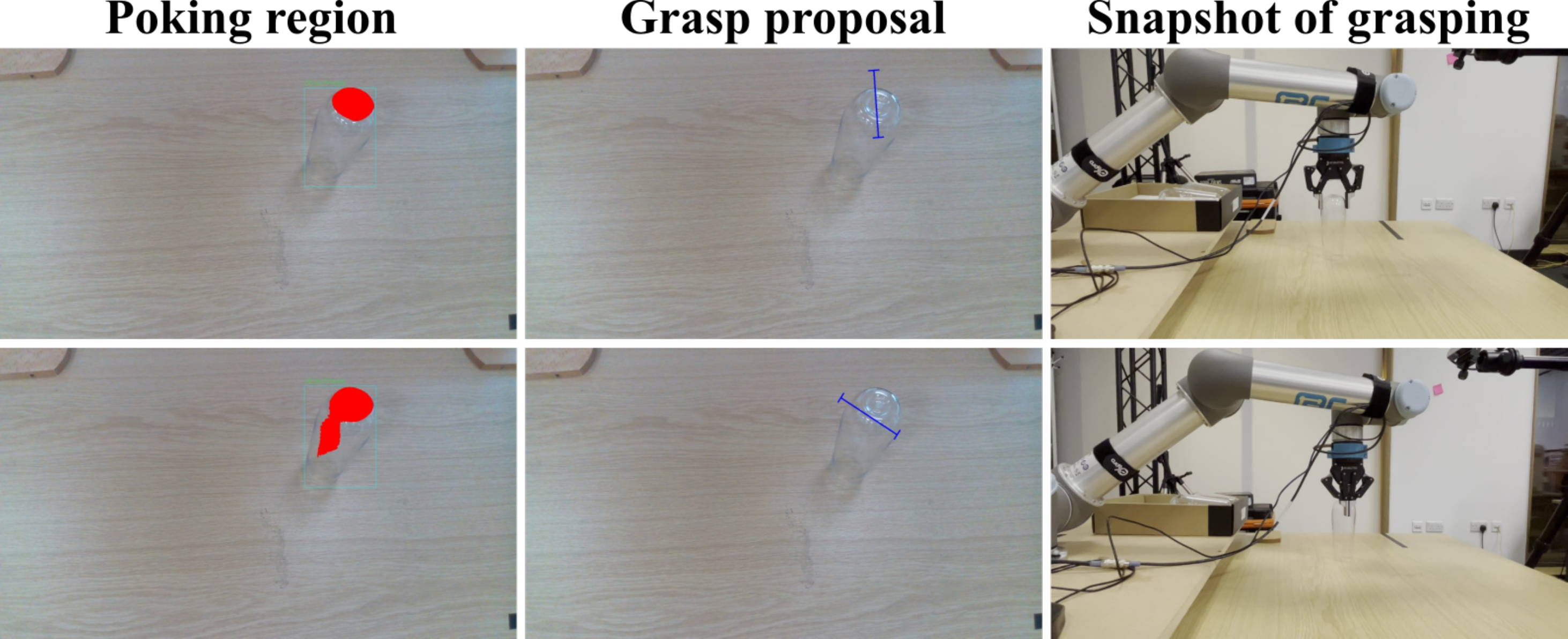}
   \caption{Examples of successful and failed grasps. \textbf{Top}: A successful grasp contributed by the poking region predicted with our PokePreNet. \textbf{Bottom}: A failure grasp caused by bad poking region segmentation when using the vanilla cross-entropy loss. The poking region, grasp proposal and snapshot of grasping are shown for each case.}
\label{fig:failuregrasp}
\end{figure}

As reported in Table~\ref{tab:re5}, the transparent objects are hard to grasp with ``Baseline1" or ``Baseline2" due to their noisy and missing depth information. Moreover, our methods significantly improve the grasping success rate from 38.9\% to 77.8\% and 85.2\% via using the accurate local profile (i.e., contact position) from vision-guided tactile poking. Similar to the results in tactile poking experiments, the bad poking region segmentation results caused by the vanilla cross-entropy loss can result in a failed grasp, i.e., the second column in Fig.~\ref{fig:failuregrasp}.

\begin{table}[t]
	\centering
		\caption{COMPARISON ON TRANSPARENT OBJECT GRASPING.}
		\label{tab:re5}
        \scalebox{1}{
		\begin{tabular}{c| c|c|c|c  }
			\hline
			 & Baseline1 &  Baseline2 & Ours1 & Ours2  \\
			\hline
			Region Type & Masks & PR & PR & PR \\
			\hline
			Loss & vanilla & PN & vanilla & PN \\
			\hline
			Localisation source & camera & camera & poking & poking \\
			\hline
			\hline
            Big disposable cup & 4/12 & 4/12 & 8/12 & 8/12 \\ \hline
            Highball cup       & 2/12 & 3/12 & 8/12 & 10/12 \\ \hline
            Rectangular cup      & 4/12 & 5/12 & 9/12 & 10/12 \\ \hline
            Vial               & 4/12 & 4/12 & 11/12 & 11/12\\\hline
            Jar                & 8/12 & 8/12 & 12/12 & 12/12\\\hline
            Mug                & 2/12 & 4/12 & 8/12 & 10/12 \\\hline
            Small disposable cup & 4/12 & 5/12 & 8/12 & 8/12 \\\hline
            Champagne cup      & 4/12 & 4/12 & 10/12 & 11/12\\\hline
            Tumble cup         & 3/12 & 5/12 & 10/12 & 12/12\\\hline
            Average success rate & 32.4\% & 38.9\% & 77.8\% & 85.2\% \\
            \hline
		\end{tabular}}
\end{table}

\subsection{Tactile Alignment for Grasping Small Objects}
Apart from providing the contact position for grasping, the tactile sensor can also sense the local shape of contact regions in the poking. Here, ``\textit{contact regions}" are the regions validated by the tactile sensor, whereas the above ``\textit{poking regions}" are from visual appearances and indicate the functional interactions of the object parts with humans or robots from the affordance perspective~\cite{nguyen2017object}.
Due to hand-eye and sensor-end-effector calibration errors mentioned in Section~\ref{Section:RobotSetup}, there would be an offset between the expected poking point predicted from vision and the centre of the contact region. The offset will result in an error in estimating the centroid of the fitted ellipse detailed in Section~\ref{Section:VisionGuidedTactilePoking} and therefore deteriorate the performance of our centroid-based grasp. To address the offset, a tactile alignment method is used to rectify the estimated centroid using the local shape obtained from the tactile image. Due to the limited perceptive field of Gelsight sensor, we only test the tactile alignment method with the small vial (the 4th object in Fig.~\ref{fig:syn_obj}). It should be noted that the tactile alignment experiment is not the main focus of this paper, but to demonstrate the potential of the current work.

\begin{figure}[t]
\centering
  \includegraphics[width=0.9\linewidth]{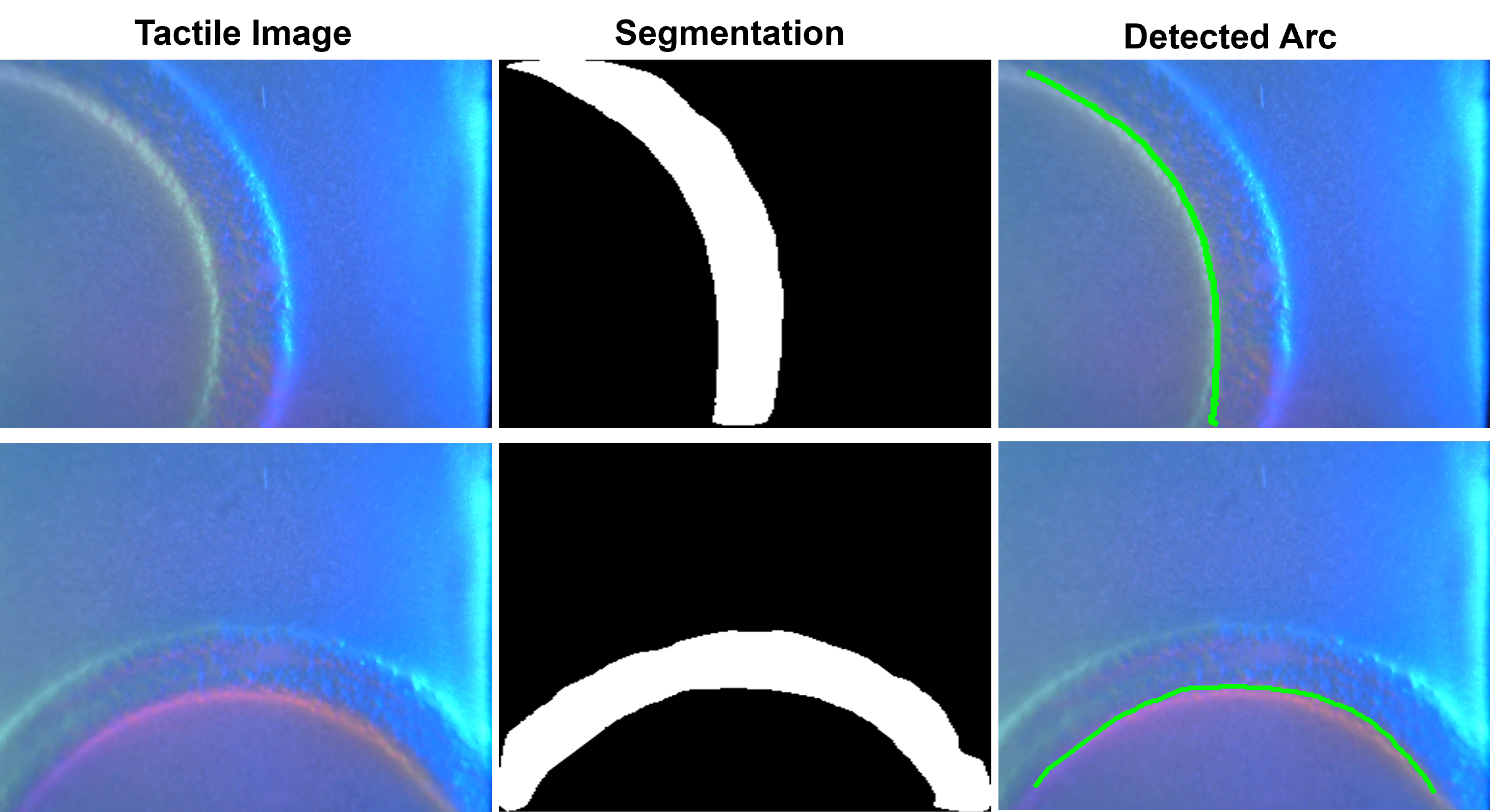}
   \caption{Examples of the position alignment predictions using tactile readings.}
\label{fig:regrasp}
\end{figure}

The contact region is first obtained with a convolutional segmengtation neural network~\cite{jiaqi2021} as shown in Fig.~\ref{fig:regrasp}. Similar to the centroid prediction in the visual images detailed in~Section~\ref{Section:VisionGuidedTactilePoking}, the OpenCV function \textit{findContours} is applied to the contact region to extract an arc of the inner ring of the vial's upper surface, and the OpenCV function \textit{fitEllipse} is used to estimate its centroid position in the tactile image frame. The pin-hole camera model~\cite{jiaqi2021} is then applied to obtain the rectified centroid position of the vial's upper surface.

To validate that the tactile alignment can enhance the robustness of grasping, we introduce a random translation error ranging from -12 $\sim$ 12 mm in $x$-axis of the world frame to the hand-eye transformation. We test with 20 attempts of grasping the vial and observe that the tactile alignment can improve the grasping success rate of the vial from 80\% to 100\%.

\subsection{Failures Analysis}

\textbf{Failures in Poking Region Segmentation.}
In the real world experiments, it has been noticed that thin poking regions of the rectangular cup (the 3rd object in Fig.~\ref{fig:syn_obj}) were hard to be detected when the object is placed upright. This was due to that the camera introduces noise to the captured images and as a result the poking regions are blurred. It could be improved by fine-tuning the PokePreNet with real dataset or incorporating it with other pixel-wise semantic segmentation methods.

\begin{figure}[t]
  \includegraphics[width=\linewidth]{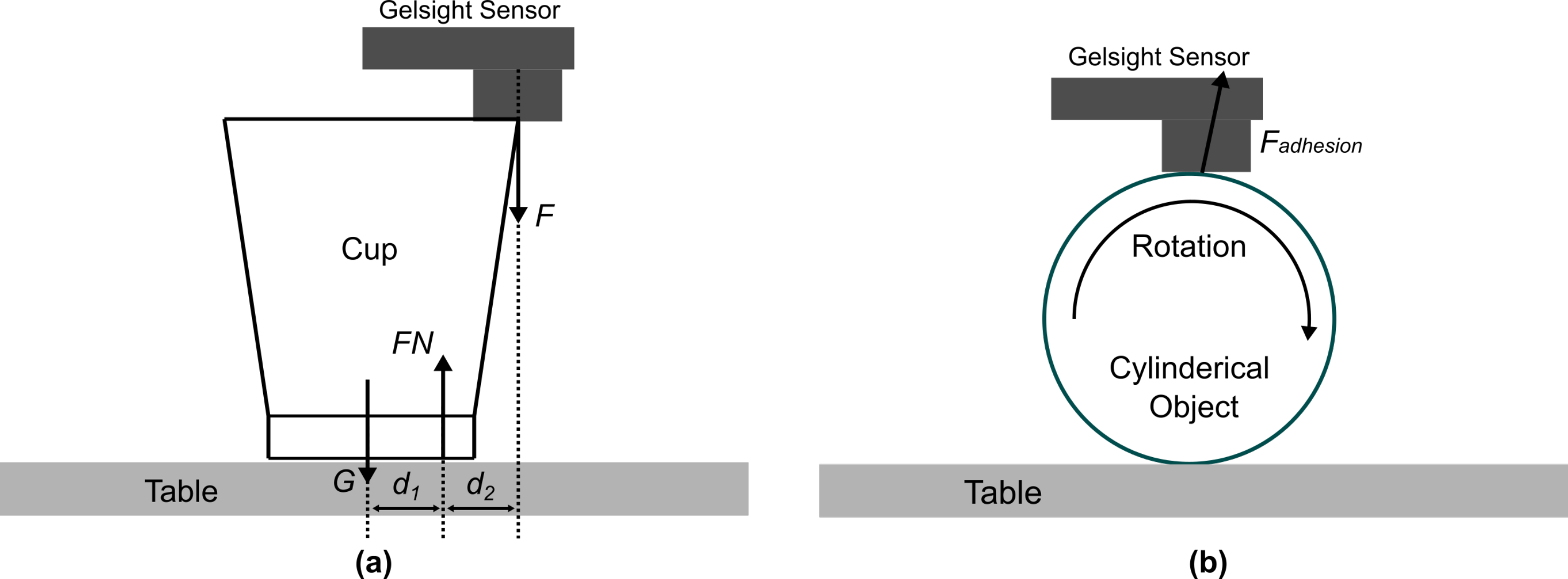}
  \caption{Side views of the tactile poking, where the GelSight sensors are in contact with (a) a cup placed upright and (b) a cylindrical object placed on its side. (a): The force analysis of the cup under static equilibrium, where $G$, $FN$ and $F$ represent the gravity of the cup, the normal force from the table and the normal force from GelSight sensor, respectively. $d_1$ and $d_2$ represent the arms of $G$ and $F$, respectively; (b): The force analysis of the cylindrical cup when an adhesion exists.}
\label{fig:failure cases}
\end{figure}

\textbf{Failures in Tactile Poking.}
One failure mode for tactile poking is the fall of transparent objects that are placed upright after being poked by the GelSight sensor. In this paper, we assume that contacting poking regions will generate reliable tactile readings, while causing minimum disturbance to the object state. However, for light objects such as disposable cups (the 1st and 8th objects in Fig.~\ref{fig:syn_obj}), the extremely small gravity cannot prevent the object from turning. As shown in Fig.~\ref{fig:failure cases}(a), when the cup is under static equilibrium, the torques of the gravity $G$ and the normal force $F$ are equal, i.e., $F*d_2 = G*d_1$. 
As a result, the maximum force applied to the disposable cup by the GelSight sensor is around $0.1N$. The robotic arm cannot react in time to such a small force due to network latency and image processing, which will lead to a failed tactile poking with the falling of the disposable cup. The excessive torque could be avoided by evenly contacting the whole poking region at the same time using a larger tactile sensor.

\textbf{Failures in Grasping.}
As shown in Tables~\ref{tab:re4} and~\ref{tab:re5}, our approach can poke and grasp cylindrical objects without making them roll on the table for most of the attempts. However, when the GelSight sensor moves away from the contacted object after poking, the adhesion between the GelSight's elastomer and the object might cause disturbance to the object's state as shown in Fig.~\ref{fig:failure cases}(b), which will lead to a failed grasp. 
We could solve this problem with a dual-arm manipulation, i.e., one arm is used to poke and fix the object on the table, and the other arm is used to explore and grasp the object.

\section{CONCLUSIONS AND DISCUSSIONS} \label{Section:6}
In this paper, we introduce a novel vision-guided tactile poking method for grasping transparent objects. Compared to previous methods, the proposed framework is the first that coordinates vision and tactile sensing to address the challenges of grasping transparent objects. The extensive experiments show that our proposed method can learn vision-guided tactile poking using only synthetic data for training and can generalise to the real world settings. The robot grasping experiments demonstrate that the informative local profile (i.e., position and local shape of contact regions) from tactile poking can enhance the performance of transparent objects grasping. 

We have the robot poke the object for once to detect the contact and update the local profiles of transparent objects, which is different from the previous tactile exploration works that contacts the object multiple times~\cite{yang2017object, watkins2019multi}. Tactile exploration can be used to estimate the object shape so as to facilitate grasping. However, in those works strong assumptions were made: either the object is 2D and the pushing process is quasi-static~\cite{yu2015shape, suresh2021tactile} or the object is fixed on the table~\cite{watkins2019multi}. These assumptions are not suitable for our cases as 3D and movable objects are used in our investigated scenarios. For example, a cylindrical cup can roll a long way on the table with a small horizontal force. In this case, the assumptions in the tactile exploration will not stand any more as the cup is highly movable and is not quasi-static. In contrast, our vision-guided tactile poking method would output a good tactile reading while maintaining minimal disturbance to the object’s state, so that the modelling of object dynamics is not needed.

The GelSight tactile sensor in this work plays two different roles. First, the GelSight sensor is used as a contact detector to validate the poking regions predicted by our PokePreNet, so as to replace the noisy depth from vision and facilitate the grasp. Second, as the GelSight sensor can extract the local shapes of small transparent objects, it is used to align grasp proposals to mitigate bias in the calibration error. When a large calibration error exists, there will be a large offset between the actual contact position and the expected position, which may lead to a failed grasping, and the geometric information of the object obtained from the tactile images can remedy grasping.

It is worth noting that the first role can also be fulfilled by other tactile sensors such as the tactile finger~\cite{piacenza2020sensorized} and the GelTip sensor~\cite{gomes2020blocks}, or force sensors like Nano 17 force/torque sensors. It means that our proposed method is general and can be easily transferred to other settings. Force sensors with sensitive force estimation could achieve better control of the poking motion compared to the GelSight sensor. However, force sensors cannot replace tactile sensors for the second role without matrix-based force readings or high-resolution tactile images. Given that, we use the tactile alignment experiment to demonstrate the advantage of using a high-resolution GelSight sensor in vision-guided tactile poking. In the future work, we will investigate the tactile alignment for grasping transparent objects further without prior knowledge of the object shape.






\bibliographystyle{IEEEtran}
\bibliography{egibib}

\end{document}